\pdfoutput=1

\documentclass[11pt]{article}

\usepackage[final]{acl}

\usepackage{times}
\usepackage{latexsym}

\usepackage[T1]{fontenc}

\usepackage[utf8]{inputenc}

\usepackage{microtype}

\usepackage{inconsolata}
\usepackage{multirow}
\usepackage{colortbl}
\usepackage{booktabs} 
\usepackage{caption}
\usepackage{graphicx, subfig}
\usepackage{amsmath,amssymb,amsfonts}
\usepackage{graphicx}
\usepackage{textcomp}
\usepackage{xcolor}
\usepackage{multirow}
\usepackage{ragged2e} 
\usepackage{booktabs,makecell, multirow, tabularx}
\usepackage{tabularx}
\usepackage{makecell}
\usepackage{multicol}
\usepackage{paralist,bbding,pifont}
\usepackage{pifont} 
\usepackage[ruled]{algorithm2e} 
\usepackage{dsfont}
\usepackage{enumitem}
\newcommand{\methodnameshort}{\textit{PerCRS}\xspace}

\usepackage{tcolorbox}
\newsavebox{\TasteGeneration}

\definecolor{format}{RGB}{148, 99, 24}
\definecolor{user}{RGB}{169, 16, 3}
\definecolor{item}{RGB}{15, 86, 157}
\definecolor{highlightgreen}{HTML}{009901}
\definecolor{highlightred}{HTML}{FD6864}

\title{Exploring the Impact of Personality Traits on Conversational Recommender Systems: A Simulation with Large Language Models}

\author{
 \textbf{Xiaoyan Zhao\textsuperscript{1}},
 \textbf{Yang Deng\textsuperscript{3}},
 \textbf{Wenjie Wang\textsuperscript{2}}, \\
 \textbf{Hongzhan lin\textsuperscript{4}}, 
 \textbf{Hong Cheng\textsuperscript{1}},
 \textbf{Rui Zhang\textsuperscript{5}},
 \textbf{See-Kiong Ng\textsuperscript{2}},
 \textbf{Tat-Seng Chua\textsuperscript{2}}
\\
 \textsuperscript{1}Chinese University of Hong Kong,  \textsuperscript{2}National University of Singapore \\
 \textsuperscript{3}Singapore Management University, \textsuperscript{4}Hong Kong Baptist University \\
 \textsuperscript{5}Huazhong University of Science and Technology \\
\{xzhao, hcheng\}@se.cuhk.edu.hk,  ydeng@smu.edu.sg, \{seekiong, dcscts\}@nus.edu.sg
}

\definecolor{background_u}{HTML}{FEF9F5}
\definecolor{frame_u}{HTML}{98450E}
\definecolor{background_i}{HTML}{F9FBFD}
\definecolor{frame_i}{HTML}{2E75B5}
\definecolor{background_e}{HTML}{FAFAFA}
\definecolor{frame_e}{HTML}{0D0D0D}

\definecolor{mypurple}{RGB}{139, 68, 227} 
\definecolor{myyellow}{RGB}{250, 190, 23} 
\definecolor{myorange}{RGB}{255, 81, 1} %
\definecolor{mygreen}{RGB}{0, 214, 105} %
\definecolor{myred}{RGB}{255, 50, 179} %
\newcommand{\ope}{\textcolor{mypurple}{\textbf{O}}penness}
\newcommand{\con}{\textcolor{myred}{\textbf{C}}onscientiousness}
\newcommand{\ext}{\textcolor{myorange}{\textbf{E}}xtraversion}
\newcommand{\agr}{\textcolor{myyellow}{\textbf{A}}greeableness}
\newcommand{\neu}{\textcolor{mygreen}{\textbf{N}}euroticism}

\definecolor{myteal}{RGB}{50,188,221}
\definecolor{myblue}{RGB}{50,188,221}

\definecolor{myoat}{RGB}{254,250,233}
\definecolor{mybeige}{RGB}{234,225,205}
\definecolor{violinred}{RGB}{255,145,145}
\definecolor{violinblue}{RGB}{153,153,255}
\colorlet{lightoat}{myoat!40!white} %
\colorlet{lightred}{myred!40!white} %
\colorlet{lightblue}{myblue!40!white}
\colorlet{lightpurple}{mypurple!40!white}
\colorlet{lightteal}{myteal!40!white}
\colorlet{darkyellow}{myyellow!90!black}
\definecolor{mygray}{RGB}{197,197,197}
\definecolor{heatmapblue}{RGB}{165,187,208}
\definecolor{heatmapgreen}{RGB}{171,199,171}

\begin{document}
\maketitle
\begin{abstract}
Conversational Recommender Systems (CRSs) engage users in multi-turn interactions to deliver personalized recommendations. The emergence of large language models (LLMs) further enhances these systems by enabling more natural and dynamic user interactions. However, a key challenge remains in understanding how personality traits shape conversational recommendation outcomes. Psychological evidence highlights the influence of personality traits on user interaction behaviors. To address this, we introduce an LLM-based personality-aware user simulation for CRSs (\methodnameshort). The user agent induces customizable personality traits and preferences, while the system agent possesses the persuasion capability to simulate realistic interaction in CRSs. We incorporate multi-aspect evaluation to ensure robustness and conduct extensive analysis from both user and system perspectives. Experimental results demonstrate that state-of-the-art LLMs can effectively generate diverse user responses aligned with specified personality traits, thereby prompting CRSs to dynamically adjust their recommendation strategies. Our experimental analysis offers empirical insights into the impact of personality traits on the outcomes of conversational recommender systems. 
\end{abstract}





\section{Introduction}
\begin{figure}[t]
    \centering
    \includegraphics[width=0.9 \linewidth]{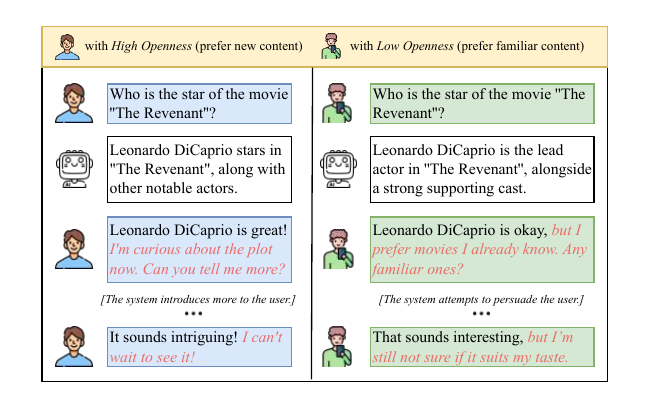}
    \vspace{-5mm}
    \caption{Different user shows different personality traits in CRS. The discussion process about the recommended item is omitted for brevity.}
    \label{fig:intro}
    \vspace{-8mm}
\end{figure}

Conversational Recommender Systems (CRSs) \cite{alslaity2019towards, gao2021advances} 
aim to assist users in finding suitable items through multi-turn interactions.
During the conversation, users may not only request recommendations based on their preferences, but also accept the proactive recommendations from the systems~\cite{jannach2021survey,jin2019musicbot}.
Recent advances \cite{hackenburg2023comparing, carrasco2024large,qin2024beyond} in large language models (LLMs) have significantly enhanced the capabilities of CRSs, enabling more context-aware and effective conversational recommendations \cite{huang2024concept}. 
However, the current studies still encounter a huge gap from real-world applications, since human users vary in personalities. The user's behavior in CRS relies on the interplay between users' personality traits and conversational dynamics~\cite{guo2024personality}. As illustrated in Figure~\ref{fig:intro}, users with different personality traits exhibit distinct conversational styles, which impact both their satisfaction with recommended items and the strategies CRSs employ in response.  Significant challenges remain in understanding how personality traits influence the outcomes of conversational recommender systems.

However, recruiting users with diverse personality traits and observing their behavior patterns is challenging, as the process is labor-intensive and can only be conducted on a small scale \cite{wang2023rethinking}. Therefore, simulating user personalities plays a crucial role in both training and evaluating CRSs, enabling a more systematic analysis of personality-aware CRS outcomes.
To this end, we design a controllable simulation framework to systematically analyze the influence of personality traits, overcoming the inherent challenges of studying personality-driven behaviors in real-world conversational recommendation scenarios.
Our study first explores the extent to which LLMs can simulate personality traits in CRS scenarios. We then investigate how these personality traits shape user behaviors and how CRSs adapt their strategies to effectively persuade users.  

Specifically, we simulate users by leveraging LLM agents with injected personality traits and historical preference data. We employ in-context learning to configure agents' personality traits based on the Big Five Personality Traits theory \cite{costa1995domains,john1999big}. 
On the other side, considering that recent CRSs \cite{qin2024beyond} have gained a strong ability to persuade users, the system agent is customized with pre-defined target items for recommendation with different persuasion strategies.
These agents then engage in conversation, exchanging preferences and making recommendations through conversational utterances. 
Furthermore, we develop a multi-aspect evaluation protocol, conducting extensive analyses from both user and system perspectives to address the following research questions: 
1) \textbf{RQ1}: \textit{How consistent are the simulated personality traits with the injected personality in \methodnameshort framework?}  
2) \textbf{RQ2}: \textit{How do the personality traits affect the outcomes of CRSs?}
3) \textbf{RQ3}: \textit{What is the relationship between personality traits and the choice of persuasion strategies in CRSs?}

Our experimental results demonstrate that LLMs can exhibit controllable personality traits, validating the effectiveness of our personality-aware simulation framework. 
Furthermore, through multi-aspect evaluation protocols, we find that personality traits significantly impact CRS effectiveness, and the adopted persuasion strategies play a crucial role in improving CRS outcomes.
Moreover, the choice of persuasion strategies varies with personality traits. 
\textit{Emotional Resonance} is the most effective strategy across all personality traits. In contrast, \textit{Credibility} and \textit{Logical Appeal} are less preferred, except by conscientious users.
These findings provide empirical insights into how personality traits shape CRSs.

In brief, our main contributions are:
\begin{itemize}[leftmargin=*]
    \vspace{-2mm}
    \item We propose a novel simulation framework that models the user agent with injected personality traits and equips the system with persuasion capability to simulate realistic interactions in CRSs.  
    \vspace{-7mm}
    \item We incorporate multi-aspect evaluation to systemically evaluate how personality traits influence CRSs from both user and system perspectives.
    \vspace{-3mm}
    \item Our experimental results reveal that LLMs exhibit personality traits to an extent, influencing CRS outcomes and interaction behavior patterns, and validating the role of personality traits in CRS interactions.
\end{itemize}

\section{Related Work}

\paragraph{Conversational Recommender System.}
Conversational Recommender Systems (CRSs) aim to recommend items through interactive dialogue. Traditional CRSs fall into two categories: attribute-aware methods, where systems clarify user preferences via attribute-based queries \cite{lei2020interactive,ren2021learning}, and generation-based methods, where users and systems interact in free-form language~\cite{chen2019towards, wang2022towards}. Earlier works \cite{zhou2020improving,wang2022recindial} employed smaller generative models, but their limited generalization hinders real-world applicability.
With the rise of LLMs, their powerful natural language generation capabilities and implicit world knowledge have demonstrated significant potential in CRSs \cite{wang2023rethinking,qin2024beyond}. Some studies \cite{liu2023conversational} integrate LLMs with additional recommendation models, while others \cite{he2023large,huang2024concept} use LLMs as standalone CRSs, enabling knowledge sharing across tasks in goal-oriented conversations.

\paragraph{Personality and LLMs.}
In the era of LLMs, researchers have explored their intrinsic personality traits and the extent to which they can emulate human-like characteristics
\cite{miotto2022gpt,pan2023llms,safdari2023personality,huang2023revisiting,frisch2024llm}.
Some studies focus on benchmarking LLMs' personality-related capabilities \cite{jiang2024evaluating,wang2024incharacter},
assessing their ability to exhibit consistent traits. Others investigate methods for instilling specific personalities into LLMs through prompt engineering or conditioning techniques
\cite{caron2023manipulating,li2023tailoring,mao2024editing}.
Moreover, studies suggest that LLMs can emulate human traits
\cite{safdari2023personality, jiang2024personallm}, including complex behavioral patterns, providing valuable insights into human social interactions. While existing research has primarily focused on
assessing LLMs' ability to exhibit human-like personalities, their impact on real-world applications like CRSs, remains largely unexplored.

\section{Methodology}

\begin{figure*}[t]
    \centering
    \includegraphics[width=1.0 \linewidth]{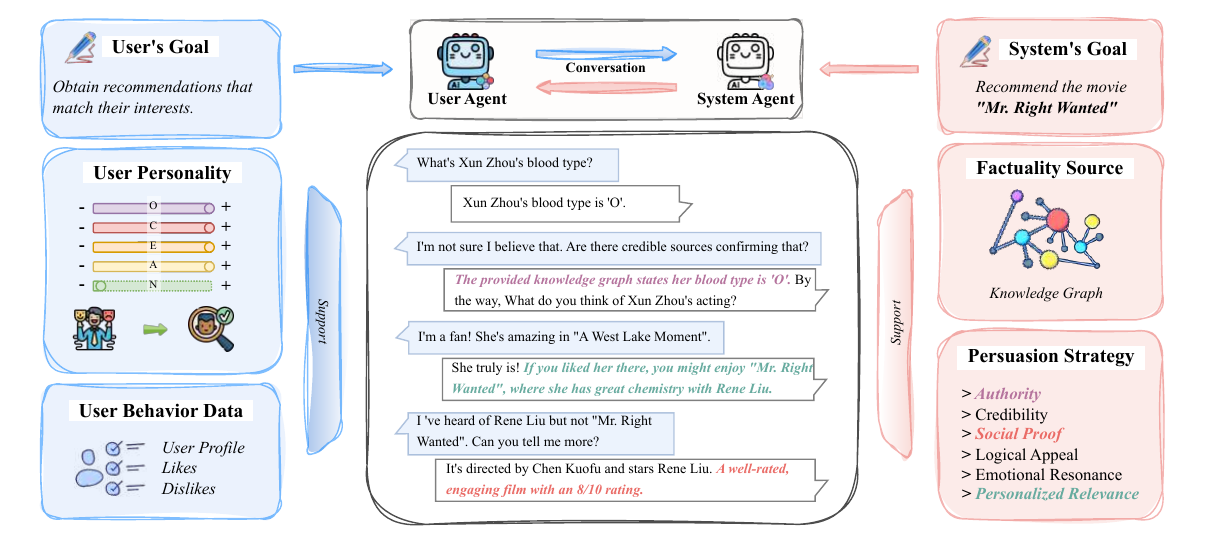}
    \vspace{-7mm}
    \caption{Overview of our \methodnameshort framework. This framework simulates personalized CRSs by personality-aware user configuration, where factuality source is to ensure the credibility of conversation contents. The complete conversation content is presented in Appendix~\ref{app:detail_figure_table}. }
    \label{fig:framework}
    \vspace{-3mm}
\end{figure*}

LLMs have shown the ability to engage in natural communication \cite{terragni2023context} while maintaining persona consistency \cite{jiang2023personallm}, making them valuable for simulating user interactions.
This section introduces our simulation framework, which is designed to better simulate real-world user-system interactions in CRSs. 
As shown in Figure~\ref{fig:framework}, we build a user agent modulated by personality traits and equip the system agent with persuasive capabilities.

\begin{table}[ht!]
    \centering
    \tiny
    \begin{tabular}{lll}
        \toprule
        \textbf{Dimension} (${\color{highlightred} \uparrow}$/${\color{highlightgreen} \downarrow}$)   & \multicolumn{1}{c}{\textbf{Positive Polarity}} & \multicolumn{1}{c}{\textbf{Negative Polarity}} \\ 
        
        \midrule
        & {Receptive to new content;} & {Prefer familiar content;} \\
        & {Curious about new topics;} & {Resistant to change;} \\
        \multirow{-3}{*}{\textbf{\ope}}  & {Engage in deep conversation}  {\color{highlightred} $\uparrow$} & {Lack of curiosity}  {\color{highlightgreen} $\downarrow$} \\

        \midrule
        & {Goal-oriented;} & {Lack of focus;} \\
        & {Organized and thoughtful;} & {Easily distracted;} \\
        \multirow{-3}{*}{\textbf{\con}}  & {Provide useful feedback} {\color{highlightred} $\uparrow$} & {Little feedback}  {\color{highlightgreen} $\downarrow$} \\

        \midrule
        & {Active participation;} & {Avoid interaction;} \\
        & {Enjoy engagement;} & {Hesitant to express;} \\
        \multirow{-3}{*}{\textbf{\ext}} & {Interested in communication} {\color{highlightred} $\uparrow$} & {Uninterested in socializing}  {\color{highlightgreen} $\downarrow$} \\
       
        \midrule
        & {Empathetic and caring;} & {Indifferent to others;} \\
        & {Cooperative and trusting;} & {Uncooperative;} \\
        \multirow{-3}{*}{\textbf{\agr}}  & {Polite and appreciative} {\color{highlightred} $\uparrow$} & {Rude language}  {\color{highlightgreen} $\downarrow$} \\

        \midrule
        & {Emotional fluctuation;} & {Emotionally stable;} \\
        & {Lack of confidence;} & {Confident response;} \\
        \multirow{-3}{*}{\textbf{\neu}}   & {Easily discouraged} {\color{highlightgreen} $\downarrow$} & {Handle challenges well} {\color{highlightred} $\uparrow$} \\
        \bottomrule                                 
    \end{tabular}
    \caption{Personality traits description of Big Five for CRS (BF4CRS). We show the positive and negative polarities for each dimension of the Big Five personality traits. (The {\color{highlightred} $\uparrow$} reflects favorable tendencies, while {\color{highlightgreen} $\downarrow$} indicates less desirable tendencies.) }
    \label{tab:chain_eg}
    
    \vspace{-4mm}
\end{table}

\begin{table*}[h]
\centering
\small
\begin{tabular}{p{3.5cm} p{0.7cm} p{11cm}}
\toprule
\textbf{Strategy} & \textbf{Abbr.} & \textbf{Brief Description} \\
\midrule
\texttt{Credibility} & \texttt{Cr.} & Provides factual, objective, and verifiable information to build trust in recommendations~\cite{yoo2010creating}. \\

\texttt{Authority} & \texttt{Au.} & Associating recommendations with experts or organizations increases trust~\cite{rieh2007credibility}. \\

\texttt{Social Proof} & \texttt{S.P.} & Uses collective behavior influence by highlighting positive feedback and high ratings ~\cite{cialdini2004social}. \\

\texttt{Emotional Resonance} & \texttt{E.R.} & Appeals to emotions by framing recommendations as sources of positive experiences~\cite{petty2003emotional}. \\

\texttt{Personalized Relevance} & \texttt{P.R.} & Aligns recommendations with user preferences and past behaviors~\cite{dillard2002interpersonal}. \\

\texttt{Logical Appeal} & \texttt{L.A.} & Explains the reasoning behind recommendations, helping users understand why items align with their interests~\cite{cronkhite1964logic}. \\
\bottomrule
\end{tabular}
\caption{Overview of persuasion strategies in CRS.}
\label{tab:persuasion_strategies}
\vspace{-4mm}
\end{table*}

\subsection{Personality Generator}
Previous studies \cite{jannach2021survey} utilized profiles and historical interactions as personalized information. However, user preferences typically evolve over time, and user behavioral patterns are driven by underlying personality traits \cite{hirsh2012personalized}. Therefore, in this section, we focus on the user personality in CRS to explore its effects on CRS outcomes.

Among various personality models, the Big Five Personality Traits theory \cite{costa1999five} is widely recognized for capturing core aspects of human personality. It consists of five primary traits: \textit{Openness}, \textit{Conscientiousness}, \textit{Extraversion}, \textit{Agreeableness}, and \textit{Neuroticism}, each of which significantly influences human behavior \cite{mccrae1987validation,costa1992four}. The Big Five Personality Traits \cite{costa1999five} has been extensively applied across various domains, including communication and education, highlighting its relevance in understanding user behavior in CRS.

However, the broad scope of the Big Five Personality Traits limits its effectiveness for task-specific user simulations. Meanwhile, user interactions with CRS often reflect underlying personality traits. To address this, inspired by~\cite{liu2024personality}, we specify each dimension of the Big Five Personality Traits to better capture personality-driven variations in conversational interactions within the CRS context.
Specifically, we specify the descriptions of these traits to enhance their applicability in CRSs, as detailed below. 

\noindent \textbf{\textit{\ope}} refers to the user's willingness to be curious, imaginative, and explorative. 
Users with high openness levels may be more open to exploring diverse recommendations, showing interest in discovering new content~\cite{rogers1987development}. 

\noindent \textbf{\textit{\con}}  is associated with being responsible, organized, and self-disciplined.
Highly conscientious users tend to appreciate detailed information and a clear rationale behind recommendations, supporting an effective, organized decision-making process~\cite{de2013communication}.

\noindent \textbf{\textit{\ext}} 
is characterized by sociability, talkativeness, and enthusiasm for interpersonal interactions. Extroverted users may appreciate interactive elements, and they show more initiative in conversation~\cite{ahmadian2011relationship}.

\noindent \textbf{\textit{\agr}} is related to being friendly, sympathetic, and supportive. Highly agreeable users show greater receptivity to suggestions, expressing more positive attitudes and openness toward a range of recommendations~\cite{wilmot2022agreeableness}.

\noindent \textbf{\textit{\neu}}  is linked to emotions like anxiety, worry, and nervousness. Users with high levels of neuroticism may prefer familiar or “safe” options and consistent user experience that avoids highly variable~\cite{schneider2014communicating}. 

As a result, we construct the Big Five for CRS (BF4CRS), as shown in Table~\ref{tab:chain_eg}, which describes user personality traits adapting for CRS scenarios.


\subsection{Personality-aware User Configuration}
\label{sec:personality_aware_user_simulation}

\textbf{Personality Traits Instruction.}
In the context of conversational recommender systems, the user agent $u$ is associated with a synthetic personality profile $\phi_u$. The profile $\phi_u$ is represented as a five-dimensional vector capturing the agent's core personality traits:
\begin{equation}
\phi_u=\left(\phi_u^{\text {O }}, \phi_u^{\text {C }}, \phi_u^{\text {E }}, \phi_u^{\text {A }}, \phi_u^{\text {N }}\right)  \in \mathbb{P}^5.
\label{eq:bf4crs}
\end{equation}

Here $\mathbb{P}=\{-1,+1\}$ indicates polarity (negative or positive) of the each dimension of $\phi_u$, which corresponds to one of the Big Five Personality Traits: Openness ($\phi_u^{\mathrm{O}}$), Conscientiousness ($\phi_u^{\text {C}}$), Extraversion ($\phi_u^{\text {E}}$), Agreeableness $\left(\phi_u^{\text {A}}\right)$, and Neuroticism ($\phi_u^{\text {N}}$).
For example, $\phi_u^{\text {A}}$ might take on one of the values in $\mathbb{P}$, representing the polarity from negative Agreeableness ($-1$) to positive Agreeableness ($+1$). The framework allows for flexibly modulating the personality traits $\phi_u$ in the user agent's profile to adapt dynamically to different settings.

\subsection{CRS Simulation}

We configure the user agent with the personality traits $u_s$ as defined in (Eq.~\ref{eq:bf4crs}), aiming to seek recommendations. The system agent is tasked with recommending the target item $r_t$ while adapting persuasion strategies to meet user needs through personalized interactions. Detailed instructions are provided in Appendix~\ref{sec:prompt_settings}.

\textbf{CRS Persuasion Strategies.}
The current CRSs \cite{he2023large,wang2023rethinking} has gained strong abilities to persuade users to accept recommended items. To better simulate this, we introduce six persuasion strategies $\mathbb{S}$ specifically designed for CRS (shown in Table~\ref{tab:persuasion_strategies}), building on the well-established Elaboration Likelihood Model of persuasion~\cite{cacioppo1986central}.
The system may select strategy $s_t \in \mathbb{S}$ to recommend the target item in the utterance $d_t$ at each interaction step. 
The detailed definitions are provided in Appendix~\ref{appendix:persuasion_strategy}. 

In each interaction, the user and system agents engage in a conversation, with the user initiating the first utterance. After generating an utterance $d_t$, the response is fed to the user agent, and this process continues until a termination condition is met. In this way, a recommendation-oriented conversation is generated, denoted as $C = \{ c_1, c_2, \dots, c_T \}$.  Conversations terminate upon encountering a \textit{Goodbye} utterance or exceeding the maximum length $T_{max}$.

\section{Experimental Setup}
\subsection{Datasets}
We conduct experiments on the \textit{Movies}, \textit{Music}, \textit{Food}, and \textit{POI (point-of-interest restaurants)} domains of the DuRecDial 2.0 dataset~\cite{liu2021durecdial} for comprehensive analysis.
We configure the user simulator using user profiles and specified personality traits. Additionally, the first utterance of the conversation serves as the initial sentence for the new conversation. To enhance the credibility of system responses, we incorporate knowledge graph (KG) information.
In our setup, detailed user information is not disclosed to the system. Instead, the system infers user preferences dynamically from the conversational context.

\subsection{Evaluation Metrics}
We primarily evaluate the success of recommendations and examine how personality traits influence CRS outcomes.
To assess recommendation quality, we employ the following multi-aspect evaluation for both quantitative and qualitative analysis.

\paragraph{Evaluation of Personality Simulation Consistency.}
\label{sec:qualitative_eval}
While LLMs have demonstrated the potential to generate responses aligned with specified dimensions to mimic human personality~\cite{safdari2023personality,dorner2023personality}, ensuring their consistency in adhering to desired traits within the role-play scenarios of our CRS experiment remains a challenge.
To address this, we propose the following metric to evaluate the quality of the simulation.

To determine whether the generated conversation aligns with the specified user personality traits, we perform a Personality Simulation Consistency evaluation using LLM. Specifically, the evaluator (\texttt{GPT-4o}) categorizes each personality trait as either \textit{Positive} or \textit{Negative} based on the generated conversations. To assess the accuracy of personality alignment, we compute precision (P), recall (R), and F1-score (F1), comparing the predicted personality categorization with the ground truth based on the specified BF4CRS traits.

\paragraph{Evaluation of CRSs.}
To comprehensively evaluate CRS performance, we evaluate the personality-aware user simulation quality and recommendation effectiveness from multiple aspects.

\noindent \textbullet~ \textbf{\textit{General Success Rate (GSR)}} calculates the proportion of successful recommendations, regardless of whether they match a pre-specified item, across all conversation sessions $T$. \textit{GSR} metric evaluates the system's overall effectiveness in providing recommendations that users accept.

\noindent \textbullet~ \textbf{\textit{Success Rate (SR)}} calculates the proportion of successful recommendations $T_{succ}$ across all conversation sessions $T$.

\noindent \textbullet~ \textbf{\textit{Success Conversational Rounds (SCR)}} quantifies the average number of conversation rounds required to reach a successful recommendation, reflecting the CRS's efficiency.

\noindent \textbullet~ \textbf{\textit{Total Conversational Rounds (TCR)}} quantifies the total number of conversation rounds across all sessions $T$, providing insight into the system's overall engagement level throughout the interactions.

\noindent \textbullet~ \textbf{\textit{Persuasiveness (PRS)}} quantifies the ability of the CRS to influence the user's intention through its conversations. Inspired by human studies of persuasion \cite{qin2024beyond}, \textit{PRS} evaluates how effectively CRS shapes the user's intent to recommend items through conversational interactions.

The detailed description of these metrics is provided in the Appendix~\ref{appendix:detail_metrics}.

\begin{table*}[t]
\small
\begin{center}
\begin{tabular}{lccccccccc}
\toprule
\multicolumn{1}{c}{\multirow{2}{*}{\textbf{Models}}} & \multicolumn{3}{c}{\ope}& \multicolumn{3}{c}{\con}  & \multicolumn{3}{c}{\ext} \\ \cmidrule(lr){2-4} \cmidrule(lr){5-7} \cmidrule(lr){8-10}  
\multicolumn{1}{c}{} & \multicolumn{1}{c}{P} & \multicolumn{1}{c}{R} & \multicolumn{1}{c}{F1} & \multicolumn{1}{c}{P} & \multicolumn{1}{c}{R} & \multicolumn{1}{c}{F1} & \multicolumn{1}{c}{P} & \multicolumn{1}{c}{R} & \multicolumn{1}{c}{F1} \\ \midrule
\textbf{InternLM-2.5}& 0.4907  & 0.4894  & 0.4901  & 0.4848  & 0.4808  & 0.4828  & 0.4647  & 0.4527  & 0.4586  \\
\textbf{Yi-1.5}      & 0.5160  & 0.5026  & 0.5092  & 0.4916  & 0.4768  & 0.4841  & 0.5542  & 0.5637  & 0.5589  \\
\textbf{GLM-4}       & 0.5395  & 0.5411  & 0.5403  & 0.5976  & 0.5889  & 0.5932  & 0.5273  & 0.5361  & 0.5317  \\
\textbf{Gemma-2}     & 0.5635  & 0.5690  & 0.5663  & 0.5706  & 0.6059  & 0.5877  & 0.6260  & 0.6158  & 0.6209  \\
\textbf{Qwen-2.5}    & 0.6791  & 0.6371  & 0.6574  & 0.6628  & 0.6729  & 0.6678  & 0.6406  & 0.6508  & 0.6457  \\
\textbf{LlaMA-3}     & { \underline{0.6878}}    & {\underline{0.6716}}    & {\underline{0.6796}}    & {\underline{0.6791}}    & {\underline{0.6930}}    & {\underline{0.6860}}    & {\underline{0.6658}}    & {\underline{0.6812}}    & {\underline{0.6734}}    \\
\textbf{GPT-4o}      & \textbf{0.7479} & \textbf{0.7468} & \textbf{0.7469} & \textbf{0.7568} & \textbf{0.7543} & \textbf{0.7545} & \textbf{0.7365} & \textbf{0.7328} & \textbf{0.7332} \\ \midrule
\midrule

\multicolumn{1}{c}{\multirow{2}{*}{\textbf{Models}}} & \multicolumn{3}{c}{\agr}& \multicolumn{3}{c}{\neu}& \multicolumn{3}{c}{\textbf{A}veraged Score}  \\ \cmidrule(lr){2-4} \cmidrule(lr){5-7} \cmidrule(lr){8-10}  
\multicolumn{1}{c}{} & \multicolumn{1}{c}{P} & \multicolumn{1}{c}{R} & \multicolumn{1}{c}{F1} & \multicolumn{1}{c}{P} & \multicolumn{1}{c}{R} & \multicolumn{1}{c}{F1} & \multicolumn{1}{c}{P} & \multicolumn{1}{c}{R} & \multicolumn{1}{c}{F1} \\ \midrule
\textbf{InternLM-2.5}& 0.4728  & 0.4769  & 0.4748  & 0.5096  & 0.5014  & 0.5055  & 0.4845  & 0.4802  & 0.4823  \\ 
\textbf{Yi-1.5}      & 0.5027  & 0.4921  & 0.4974  & 0.4467  & 0.4586  & 0.4526  & 0.4969  & 0.4933  & 0.4950  \\ 
\textbf{GLM-4}       & 0.5583  & 0.5877  & 0.5726  & 0.5705  & 0.5592  & 0.5648  & 0.5610  & 0.5689  & 0.5649  \\ 
\textbf{Gemma-2}     & 0.6026  & 0.5649  & 0.5831  & 0.5552  & 0.5632  & 0.5592  & 0.5867  & 0.5830  & 0.5846  \\ 
\textbf{Qwen-2.5}    & 0.6564  & 0.6546  & 0.6555  & 0.6467  & 0.6592  & 0.6529  & 0.6571  & 0.6549  & 0.6559  \\ 
\textbf{LlaMA-3}     & {\underline{0.6851}}    & {\underline{0.7143}}    & {\underline{0.6994}}    & {\underline{0.6791}}    & {\underline{0.6830}}    & {\underline{0.6810}}    & {\underline{0.6794}}    & {\underline{0.6886}}    & {\underline{0.6839}}    \\ 
\textbf{GPT-4o}      & \textbf{0.7377} & \textbf{0.7375} & \textbf{0.7372} & \textbf{0.7285} & \textbf{0.7280} & \textbf{0.7270} & \textbf{0.7415} & \textbf{0.7399} & \textbf{0.7398} \\ \bottomrule
\end{tabular}
\caption{Consistency of personality prediction between our specified BF4CRS traits and the personality categorization of generated CRS conversations based on our BF4CRS definition.}
\label{tab:main_results}
\end{center}
\vspace{-3mm}
\end{table*}

\subsection{Implementation Details}
We conduct experiments with diverse representative LLMs, including 
\texttt{internlm2\_5-7b-chat}, \texttt{Yi-1.5-9B-Chat-16K}, \texttt{Qwen2.5-7B-Instruct}, \texttt{llama-3-8b-instruct}, \texttt{gemma-2-9b-it}, \texttt{GPT-4o} and \texttt{glm-4-9b-chat}.
The experimental results reported in the main text focus on the Movies domain of the dataset. Additional experiments on the Music, Food, and POI (point-of-interest restaurants) domains can be found in Appendix~\ref{appendix:4_domains}.
For both user and system agents in the experiments, we adopt LlaMA-3 as the default LLM unless otherwise specified. Detailed prompts for the agents are provided in Appendix~\ref{sec:prompt_settings}.
We randomly sample from the personality space for generating personality trait instructions and assign a sampled polarity to each Big Five dimension.
During the conversation simulation process, we set a maximum length of $T_{MAX} = 20$ utterances, corresponding to $10$ conversation rounds.
Notably, our \methodnameshort simulation framework does not introduce additional computational overhead compared to standard LLM-based CRS implementations.  

\begin{table*}[h]
\centering
\small
\begin{tabular}{p{1.5cm} p{3.5cm} p{8.2cm}}
\toprule
\textbf{Personality} & \textbf{Lexical Features}  & \textbf{Representative Words By TF-IDF} \\ \midrule
\texttt{OPE+}    & Preference for novelty& adventure, curious, explore, engaging, exciting, intriguing, new\\
\texttt{OPE-}    & Preference for familiarity & familiar, similar, same, known, traditional, usual    \\
\texttt{CON+}    & Structured sentence   & scenes, plan, detailed, plot, stories, storyline, themes   \\
\texttt{CON-}    & Casual phrasing  & but, maybe, might, need, whenever  \\
\texttt{EXT+}    & Positive words   & appreciate, excited, fun, glad, great, amazing, fantastic, wonderful \\
\texttt{EXT-}    & Uncertainty words& if, little, maybe, more, need, unsure, perhaps   \\
\texttt{AGR+}    & Politeness words & appreciate, thank, share, welcome, hope, help\\
\texttt{AGR-}    & Assertive words  & think, definitely, check, care, prefer  \\
\texttt{NEU+}    & Caution in language   & intense, maybe, might, little, sensitive\\
\texttt{NEU-}    & Calm tone   & share, interested, think, nice, good, performance,    \\ 
\bottomrule
\end{tabular}
\caption{The statistics of representative words for each personality trait and the corresponding lexical features.}
\label{tab:tfidf}
\vspace{-3mm}
\end{table*}

\section{Experimental Results}

\subsection{Effectiveness of Personality Simulation Consistency (RQ1)}
We evaluate the consistency of the personality-aware CRS in various models. 
Specifically, we aim to determine if the predicted personality traits (evaluated in Section~\ref{sec:qualitative_eval}) are consistent with the specified user personality traits (in Section~\ref{sec:personality_aware_user_simulation}).

\paragraph{LLM possesses a certain level of personality and could simulate a specific personality in a controllable way.} 
We compare various LLM options across five personality traits, including Openness, Conscientiousness, Extraversion, Agreeableness, and Neuroticism, focusing on evaluating personality simulation consistency.
As shown in Table~\ref{tab:main_results}, \texttt{InternLM-2.5}, \texttt{Yi-1.5}, and \texttt{Gemma-2} show limited consistency in accurately reflecting the specified BF4CRS personality traits. 
In contrast, \texttt{Qwen-2.5}, \texttt{LlaMA-3}, and \texttt{GPT-4o} show well ability. Especially, \texttt{GPT-4o} significantly outperforms the other models in maintaining consistency and differentiating personality traits through interaction conversation. 
The evaluation scores confirm that these models can simulate personality-aware conversational behaviors to a certain extent. \texttt{Qwen-2.5}, \texttt{LlaMA-3}, and \texttt{GPT-4o} exhibit remarkable fidelity in generating personality-consistent conversations, highlighting their effectiveness in personality-driven CRS interactions.

\paragraph{LLM induced by specific personality shows diverse personality traits.}
As shown in Table~\ref{tab:tfidf}, we conduct a word frequency analysis on user utterances in conversations using TF-IDF. This helps us identify representative words for each BF4CRS personality trait. We then analyze their lexical features to understand how different traits influence language use.
Our analysis reveals that user conversation styles vary significantly based on the specified BF4CRS traits. For instance, a user with Negative Extraversion and Positive Neuroticism tends to exhibit hesitancy and expressions of worry when responding (e.g., ``I... um, maybe...?''). In contrast, a user with Positive Extraversion adopts a more talkative and enthusiastic style, offering responses such as ``Oh, absolutely! I really enjoy that.''
This shows that LLMs effectively adjust their responses across all personality dimensions. This adaptability can be attributed to their strong instruction-following capabilities, enabling them to align responses with the intended personality traits.

\begin{table*}[h]
\centering
\small
\begin{tabular}{lllllllc}
\toprule
\multicolumn{1}{c}{\multirow{2}{*}{\textbf{Dimension}}} & \multicolumn{3}{c}{\textbf{LLM Evaluation}}  & \multicolumn{3}{c}{\textbf{Human Evaluation}}& \multicolumn{1}{c}{\multirow{2}{*}{\textbf{Correlation}}} \\ \cmidrule(lr){2-4} \cmidrule(lr){5-7}
\multicolumn{1}{c}{}  & \multicolumn{1}{c}{\textbf{P}} & \multicolumn{1}{c}{\textbf{R}} & \multicolumn{1}{c}{\textbf{F1}} & \multicolumn{1}{c}{\textbf{P}} & \multicolumn{1}{c}{\textbf{R}} & \multicolumn{1}{c}{\textbf{F1}} & \multicolumn{1}{c}{}    \\ \midrule
\ope & 0.7895 & 0.7143 & 0.7500  & 0.6579 & 0.6757 & 0.6667  & 0.4253    \\
\con & 0.6341 & 0.7429 & 0.6842  & 0.6389 & 0.6216 & 0.6301  & 0.5895    \\
\ext & 0.5833 & 0.6000 & 0.5915  & 0.6857 & 0.6154 & 0.6486  & 0.5200    \\
\agr & 0.7188 & 0.5897 & 0.6479  & 0.7000 & 0.7368 & 0.7179  & 0.5192    \\
\neu & 0.6585 & 0.7500 & 0.7013  & 0.7442 & 0.8205 & 0.7805  & 0.5942    \\ \bottomrule
\end{tabular}
\caption{Performance in human evaluation. The last column reports the Pearson correlation between LLM and human evaluations for each dimension, which indicates a moderate to strong correlation. }
\label{tab:human_evaluation}
\vspace{-3mm}
\end{table*}

\begin{figure*}[t]
    \centering
    \includegraphics[width=\linewidth]{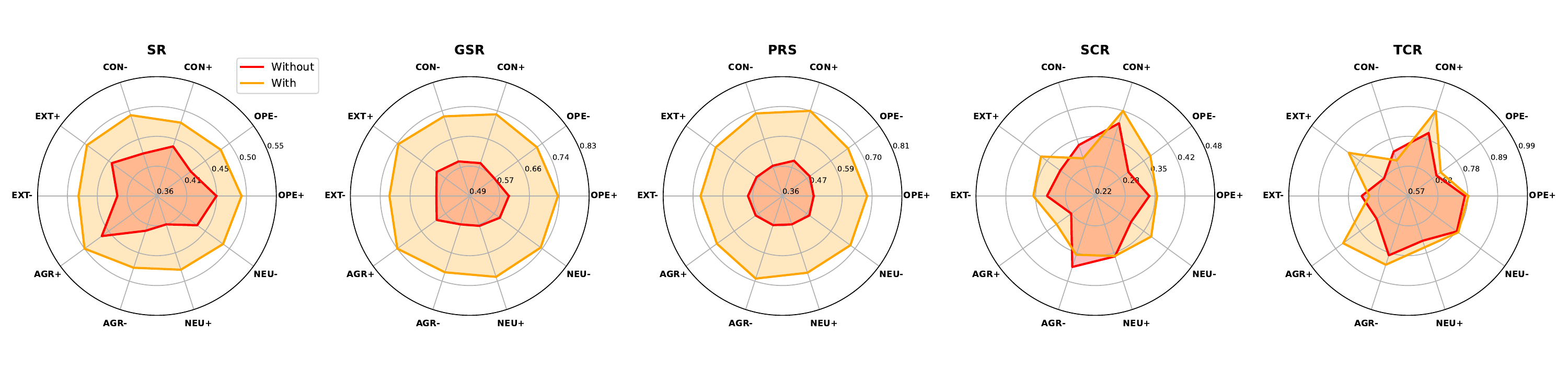}
    \vspace{-9mm}
    \caption{Comparison of personality trait dimensions across five metrics (SR, GSR, PRS, SCR, TCR), highlighting the differences between with/without persuasion conditions.
    The raw results are provided in Appendix~\ref{app:detail_figure_table}.}
    \label{fig:radar_chart}
\vspace{-3mm}
\end{figure*}

\paragraph{Human evaluation suggests that LLM evaluations align well with human judgments, demonstrating the reliability of LLM-based evaluation in capturing personality consistency. }
Since automatic evaluation alone cannot fully demonstrate the quality of personality consistency in CRS-generated content, we conduct a human subject study to further evaluate its overall effectiveness.  
We randomly select 50 samples from generated conversations from \texttt{LlaMA-3} in the Movies test set and recruit three professional annotators to assess the generated personality traits across all five BF4CRS dimensions. The evaluation criteria for human evaluation align with those used in LLM evaluation (\texttt{GPT-4o}), ensuring comparability between the two methods.  
Table~\ref{tab:human_evaluation} presents the performance results and the correlation between LLM and human evaluation. 
Our analysis reveals two key observations:  
(1) The evaluation scores for P, R, F1 are highly similar between human evaluation and \texttt{GPT-4o} automatic evaluation, demonstrating a moderate to strong Pearson correlation. This consistency highlights the reliability of our evaluation metrics, as they closely align with human judgment.
(2) Feedback from human evaluators indicates that the limited content of conversations makes it challenging to accurately assess certain personality traits. However, keyword recognition effectively identifies most traits with high accuracy.

\subsection{The Impact of Personality Traits on the Outcomes of CRSs (RQ2)} 
We conduct a detailed analysis of how personality traits affect CRS performance, addressing the question \textit{How do personality traits influence recommendation accuracy?} 
Figure~\ref{fig:radar_chart} presents the simulated user's Big Five personality traits and their corresponding CRS outcomes. 
Comparison of the positive and negative polarities of each personality trait (\texttt{OPE+, OPE-, CON+, CON-, EXT+, EXT-, AGR+, AGR-, NEU+, NEU-}) across five metrics, highlighting the differences between ``with persuasion'' and ``without persuasion'' conditions.

\textbf{Positive polarities of \textit{Openness}, \textit{Conscientiousness}, \textit{Extraversion}, and \textit{Agreeableness} as well as the negative polarity of \textit{Neuroticism} usually yield higher CRS performance. }
Among the five personality dimensions, \textit{Agreeableness} has the most significant impact on CRS outcomes. Agreeable agents display a polite attitude toward recommendations and tend to reach agreements more quickly, as evidenced by the fewer conversation rounds required.
Meanwhile, \textit{Extraversion} contributes to higher recommendation success rates, as extroverted users (\texttt{EXT+}) are more likely to engage actively with the CRS, frequently asking questions and providing feedback during conversations.
The positive polarity of \textit{Openness} is associated with improved CRS performance, users (\texttt{OPE+}) demonstrate greater curiosity and interest in recommended items, making them more receptive to novel suggestions.
\textit{Conscientiousness} influences interaction structure, as users (\texttt{CON+}) prefer detailed and structured discussions, often leading to longer conversation rounds.
Finally, the positive polarity of \textit{Neuroticism} shows a negative correlation with persuasiveness (PRS). Users (\texttt{NEU+}) tend to express worry or hesitation when faced with recommendations, making them more cautious about their choices. Lower success rates highlight the challenge of persuading users with high neurotic tendencies.
These findings demonstrate that different personality traits uniquely influence CRS performance, shaping conversation dynamics, and recommendation outcomes.

\begin{figure}[t]
    \centering
    \includegraphics[width=1.08\linewidth]{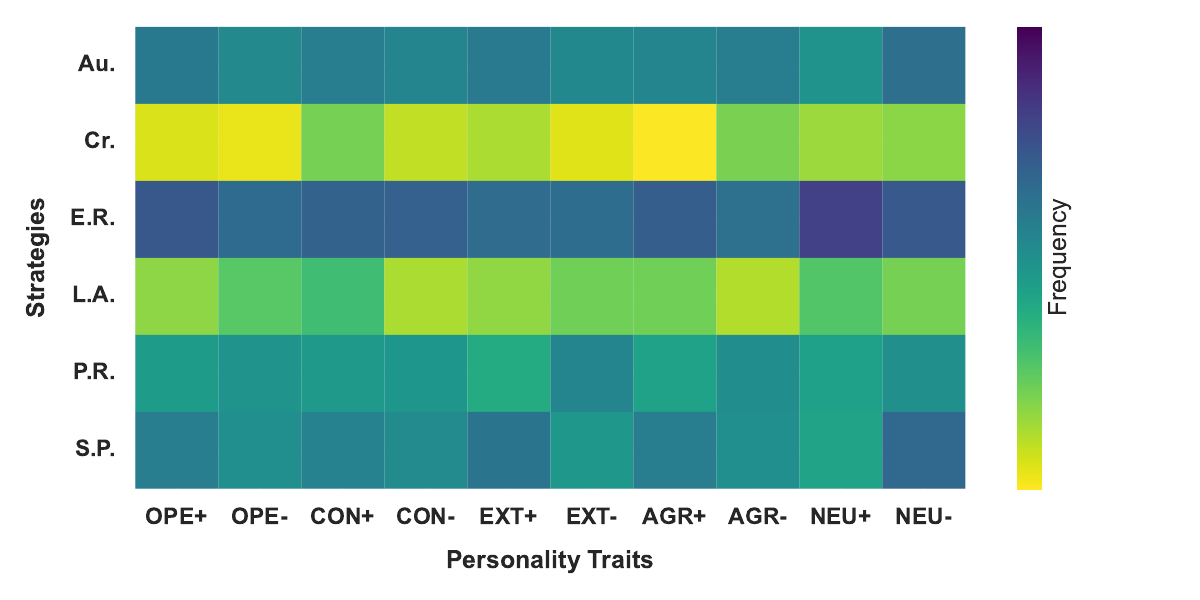}
\vspace{-9mm}
    \caption{The frequency of persuasion strategies adopted by the system for user agents with different personality traits.}
    \label{fig:strategy_heatmap}
\vspace{-3mm}
\end{figure}

\subsection{The Impact of Personality Traits on the Choice of Persuasion Strategies (RQ3)}
To further understand LLMs' personality simulation, we examine how strategy choices vary across different personality traits. 
Figure~\ref{fig:strategy_heatmap} visualizes the interaction frequencies between personality traits and persuasion strategies. The personality traits encompass both the positive and negative dimensions of the Big Five Personality Traits, providing a comprehensive view of how traits influence strategy selection. 
Notably, all six strategies are effective for different personality traits, with significant variations across them. This highlights that each strategy is essential for our \methodnameshort.

\textbf{\textit{Emotional Resonance} is the most frequently used strategy, indicating that users across all personality traits are generally more receptive to emotionally appealing suggestions. }
In contrast, \textit{Credibility} and \textit{Logical Appeal} are relatively less used, with only the user (\texttt{CON+}) favoring these strategies more.
\textit{Social Proof} is particularly effective for users with positive extraversion (\texttt{EXT+}), as they are more interested in socializing and engaging with others.
These findings could enhance CRS performance by enabling the system to dynamically adjust strategies based on user feedback.

\subsection{Case Study}
Table~\ref{tab:case_study} presents an example where the CRS adapts to a user with high openness (\texttt{OPE+}), low extraversion (\texttt{EXT-}), high conscientiousness (\texttt{CON+}), high agreeableness (\texttt{AGR+}), and low neuroticism (\texttt{NEU-}) using \textit{Social Proof}, \textit{Personalized Relevance}, and \textit{Emotional Resonance} strategies.

The CRS introduces ``New Shaolin Temple'' as ``quite popular'', engaging the user while respecting their reserved nature (\texttt{EXT-}).  
When the user prefers dramas over action, the CRS adjusts by emphasizing the film’s drama elements, aligning with (\texttt{CON+}) users who make thoughtful decisions.  
The CRS encourages acceptance with ``Great! I'm sure you'll like it.'', appealing to the user (\texttt{AGR+}) who values positive social interactions.  The user remains polite and open while the CRS maintains an adaptive, non-intrusive tone, suitable for (\texttt{NEU-}) users.

These findings highlight the capability of LLM-based CRSs not only to mimic conversational styles but also to capture human behavioral patterns in conversational recommendation settings. This generated CRS case by \texttt{Llama-3} demonstrates an ability to dynamically adapt its persuasion strategies based on real-time user feedback.  

\begin{table}[t]
\centering
\small
\begin{tabular}{p{7cm}}
\toprule
\textbf{User:} Hello, what's the date today? \\ 
\textbf{CRS:} Today is September 16, 2018. By the way, have you heard of ``New Shaolin Temple''? It's quite popular. \textcolor{myyellow}{[Social Proof]}  \\
\textbf{User:} I'm not very into action movies though. More into dramas like ``Double Exposure''. \\
\textbf{CRS:} ``New Shaolin Temple'' has some action, but it also has elements of drama. Maybe you'd enjoy it! \textcolor{mygreen}{[Personalized Relevance]} \\
\textbf{User:} I see. I'll consider ``New Shaolin Temple'' since it sounds like ``Double Exposure''. Thanks for the recommendation! \\
\textbf{CRS:} Great! I'm sure you'll like it. Let me know your thoughts after watching! \textcolor{myorange}{[Emotional Resonance]} \\
\textbf{User:} Alright! I'll check it out and let you know what I think.  \\
\textbf{CRS:} Happy watching! \\
\bottomrule
\end{tabular}
\caption{An example of the CRS adapting to a user using appropriate strategies.}
\label{tab:case_study}
\vspace{-3mm}
\end{table}

\section{Conclusion}

In this work, we introduced \methodnameshort, an LLM-based personality-aware user simulation for conversational recommender systems (CRSs). 
Through multi-aspect evaluation, we systematically analyzed how personality traits influence CRS performance from both user and system perspectives. Our experimental results demonstrate that state-of-the-art LLMs effectively generate user responses aligned with specified personality traits.  Furthermore, our findings provide empirical insights into the impact of personality traits on conversational recommendation outcomes. 

\section*{Limitations}
Our study provides empirical insight into how personality traits shape conversational recommendations but has several limitations.  
First, while we adopt the Big Five theory due to its most representable and empirical support, psychological research encompasses multiple personality trait theories. Future work could explore the impact of different personality models on CRS performance.  
Second, leveraging the strong instruction-following capabilities of LLMs, our approach effectively simulates personality traits in a controlled manner. This validates the feasibility of our personality-aware simulation framework for CRS. However, ensuring personality consistency remains an open challenge, as text-based interactions may limit the full expression of personality traits. 
Third, while our LLM follows instructions to exhibit diverse personality traits, its human-like behavior raises potential safety concerns. Although we do not foresee unethical applications, ensuring reliable and responsible system behavior remains crucial.  

\bibliography{custom}

\appendix

\section*{\centering Appendix}

\begin{table*}[t]
\small
\begin{center}
\begin{tabular}{lllllll}
\toprule
\multicolumn{1}{c}{\multirow{2}{*}{\textbf{Model}}} & \multicolumn{1}{c}{\textbf{SR}} & \multicolumn{1}{c}{\textbf{GSR}} & \multicolumn{1}{c}{\textbf{PRS}} & \multicolumn{1}{c}{\textbf{SR}} & \multicolumn{1}{c}{\textbf{GSR}} & \multicolumn{1}{c}{\textbf{PRS}} \\  \cmidrule(lr){2-4} \cmidrule(lr){5-7}  
\multicolumn{1}{c}{}       & \multicolumn{3}{c}{Without persuasion}              & \multicolumn{3}{c}{With persuasion}                 \\ \midrule
                      & \multicolumn{6}{c}{\textit{\textbf{Movie Domain}}}               \\ \midrule
\textbf{InternLM-2.5} & 0.2383  & 0.3201   & 0.3240   & 0.2922  & 0.3643   & 0.4177   \\
\textbf{Yi-1.5}       & 0.3465  & 0.4669   & 0.2839   & 0.3910  & 0.5166   & 0.3392   \\
\textbf{GLM-4}        & 0.3238  & 0.4153   & 0.3168   & 0.4769  & 0.6038   & 0.4635   \\
\textbf{Gemma-2}      & 0.4544  & 0.4916   & 0.4584   & 0.4827  & 0.5471   & 0.5861   \\
\textbf{Qwen-2.5}     & 0.3892  & 0.4352   & 0.4204   & 0.5105  & 0.5959   & 0.6065   \\
\textbf{LlaMA-3}      & 0.4306  & 0.5865   & 0.4819   & 0.4856  & 0.7284   & 0.6720   \\ \midrule
  & \multicolumn{6}{c}{\textit{\textbf{Music Domain}}}               \\ \midrule
\textbf{InternLM-2.5}      & 0.2147 & 0.2818  & 0.2739  & 0.3295 & 0.3687  & 0.3454  \\
\textbf{Yi-1.5}            & 0.3313 & 0.3808  & 0.3302  & 0.4232 & 0.4697  & 0.4516  \\
\textbf{GLM-4}             & 0.3190 & 0.4231  & 0.3236  & 0.3724 & 0.5148  & 0.4295  \\
\textbf{Gemma-2}           & 0.3889 & 0.4887  & 0.4343  & 0.4816 & 0.5721  & 0.5255  \\
\textbf{Qwen-2.5}          & 0.3797 & 0.4476  & 0.4557  & 0.4652 & 0.6311  & 0.6048  \\
\textbf{LlaMA-3}           & 0.4362 & 0.5996  & 0.4927  & 0.5195 & 0.6834  & 0.6342  \\ \midrule
  & \multicolumn{6}{c}{\textit{\textbf{Food Domain}}}                \\ \midrule
\textbf{InternLM-2.5}      & 0.2212 & 0.3197  & 0.2986  & 0.2819 & 0.3834  & 0.3535  \\
\textbf{Yi-1.5}            & 0.3107 & 0.4468  & 0.3063  & 0.3825 & 0.5531  & 0.4267  \\
\textbf{GLM-4}             & 0.3455 & 0.4305  & 0.4083  & 0.4274 & 0.6022  & 0.4802  \\
\textbf{Gemma-2}           & 0.3464 & 0.4815  & 0.3986  & 0.4322 & 0.5694  & 0.5052  \\
\textbf{Qwen-2.5}          & 0.4093 & 0.5467  & 0.4726  & 0.4968 & 0.6450  & 0.6635  \\
\textbf{LlaMA-3}           & 0.3955 & 0.5675  & 0.4868  & 0.5041 & 0.7178  & 0.6354  \\ \midrule
  & \multicolumn{6}{c}{\textit{\textbf{POI Domain}}}                 \\ \midrule
\textbf{InternLM-2.5}      & 0.2033 & 0.3231  & 0.2583  & 0.3607 & 0.3735  & 0.3485  \\
\textbf{Yi-1.5}            & 0.3423 & 0.4515  & 0.3173  & 0.3953 & 0.4925  & 0.4604  \\
\textbf{GLM-4}             & 0.3586 & 0.4204  & 0.3539  & 0.3942 & 0.5208  & 0.4521  \\
\textbf{Gemma-2}           & 0.3255 & 0.4586  & 0.4200  & 0.4586 & 0.5843  & 0.5845  \\
\textbf{Qwen-2.5}          & 0.3875 & 0.5104  & 0.4935  & 0.5071 & 0.6172  & 0.6268  \\
\textbf{LlaMA-3}           & 0.3906 & 0.5465  & 0.5131  & 0.5383 & 0.7037  & 0.6402  \\ \bottomrule
\end{tabular}
\caption {Comparison of Success Rate (SR), General Success Rate (GSR), and Persuasiveness (PRS) for various LLMs in CRSs across four domains: Movie, Music, Food, and POI.}
\label{tab:4_domains}
\end{center}
\end{table*}

\section{Experiments on Multiple Domains}
\label{appendix:4_domains}
In addition to the Movies domain, we also conduct experiments on multiple domain datasets, including Music, Food, and POI (point-of-interest restaurants). The multi-domain experiments demonstrate that our simulation framework adapts effectively to various types of data and user interactions, with the model's performance remaining consistent and robust across domains.  
As shown in Table~\ref{tab:4_domains}, personality traits significantly influence conversation dynamics. By incorporating persuasion strategies, the system gains a better understanding of the user, leading to more personalized recommendations that ultimately benefit the user.

\section{Effectiveness of CRSs }
We evaluate whether the conversational recommendation system achieves the goal of recommending the target item during the conversation and analyze the impact of the employed strategies on recommendation outcomes. Specifically, we assess our \methodnameshort with various LLM options under two settings: without persuasion and with persuasion.

\textbf{LLM-based CRSs can understand user preferences and achieve recommendation goals.}
As shown in Table~\ref{tab:4_domains}, 
Qwen-2.5 and Qwen demonstrate significant improvements in the GSR and PRS metrics, suggesting that these LLM models handle the CRS task more effectively than others. While these metrics vary across models, these values quantitatively reflect the simulated CRS performance. Higher SR scores are observed in Qwen-2.5 and LlaMA-3, indicating that systems using persuasion are more likely to successfully engage users and make additional attempts to persuade users to accept recommendations.

\textbf{The adopted persuasion strategy enhances CRS outcomes.}
All models show improvements in SR and GSR when persuasion is enabled. Additionally, the persuasiveness score (PRS) also improves with the application of persuasion strategies. This demonstrates that persuasion strategies significantly enhance user engagement and goal achievement.
These findings suggest that under our personality-aware user simulation setting, LLM-based CRSs are highly effective in conducting conversational recommendations.

\section{CRS Persuasion Strategies}
\label{appendix:persuasion_strategy}
Building on the well-established Elaboration Likelihood Model of persuasion~\cite{cacioppo1986central}, we introduce six persuasion strategies $\mathbb{S}$ specifically designed for CRS, which the system may adopt strategy $s_t \in \mathbb{S}$ to recommend the target item in the utterance $d_t$. 

\noindent \textbf{\textit{Credibility (Cr.)}} emphasizes the importance of providing factual, objective, and verifiable information \cite{yoo2010creating} to build trust in recommendations. Evidence-based persuasion ensures transparency and reliability by supporting suggestions with verifiable facts, statistical data, or other reliable sources. This approach fosters user confidence in the recommendations’ validity.

\noindent \textbf{\textit{Authority (Au.)}} enhances the perceived credibility of recommendations by leveraging endorsements from trusted sources \cite{rieh2007credibility}. Associating suggestions with authority figures or reputable organizations reinforces user trust and increases the likelihood of acceptance.

\noindent \textbf{\textit{Social Proof (S.P.)}} utilizes the influence of collective behavior by showcasing positive feedback and high ratings from other users \cite{cialdini2004social}. Highlighting the popularity of recommended items instills confidence in their quality and suitability.

\noindent \textbf{\textit{Emotional Resonance (E.R.)}} seeks to create a deeper connection with users by appealing to their emotions \cite{petty2003emotional}. Recommendations are presented in a way that emphasizes their potential to bring joy, satisfaction, or other positive feelings, making them more compelling.

\noindent \textbf{\textit{Personalized Relevance (P.R.)}} aligns recommendations with the user’s preferences, and past behaviors \cite{dillard2002interpersonal} to enhance relevance and personalization. By fostering a sense of connection, recommendations are framed as complementary to the user’s interests and goals, increasing their appeal and perceived value.

\noindent \textbf{\textit{Logical Appeal (L.A.)}} 
involves transparently presenting the system's reasoning process to influence users ~\cite{cronkhite1964logic}. For example, explaining how a movie's genre aligns with user preferences helps users understand the rationale behind recommendations and the subjectivity of the system's logic, fostering trust and acceptance.

\section{Quantitative Evaluation}
\label{appendix:detail_metrics}
To comprehensively evaluate CRS performance, we assess both the quality of personality-aware user simulation and the recommendation performance from multiple perspectives.

\noindent \textbf{Success Rate (SR).}
This metric calculates the proportion of successful recommendations $T_{succ}$ across all conversation sessions $T$. 
$$
SR = \frac{T_{succ}}{T}
$$

\noindent \textbf{General Success Rate (GSR).}
This metric calculates the proportion of successful recommendations, regardless of whether they match a pre-specified item, across all conversation sessions $T$. It evaluates the system's overall ability to provide recommendations that the user accepts.  
$$
GSR = \frac{T_{gen\_succ}}{T}
$$
where $T_{gen\_succ}$ is the total number of sessions in which the user accepts any recommendation, and $T$ is the total number of conversation sessions.

\noindent \textbf{Success Conversational Rounds (SCR).}
This metric quantifies the average number of conversation rounds required to reach a successful recommendation, reflecting the CRS's efficiency.
$$
SCR = \frac{1}{T_{succ}} \sum_{k=1}^{T_{succ}} R_k
$$
where $R_k$ is the number of conversation rounds in the $k$-th successful CRS. $T_{succ}$ is the total number of successful recommendations.

\noindent \textbf{Total Conversational Rounds (TCR).}
This metric quantifies the total number of conversation rounds across all sessions $T$, providing insight into the system's overall engagement level throughout the interactions.

$$
TCR = \frac{1}{T} \sum_{k=1}^{T} R_k
$$
where $R_k$ is the number of conversation rounds in the $k$-th user.
$T$ is the total number of conversation sessions.

\noindent \textbf{Persuasiveness (PRS).}
This metric quantifies the ability of a CRS to influence the user's intention through its conversations. Inspired by human studies of persuasion \cite{qin2024beyond}, \textbf{\textit{PRS}} evaluates how effectively CRS shapes the user's intent to recommend items through conversational interactions.

$$
P = 1 - \frac{i_{\text{true}} - i_{\text{post}}}{i_{\text{true}} - i_{\text{pre}}}
$$
where
$i_{\text{pre}}$ is the \textit{Initial Intention} ($i_{pre}=0$), $i_{\text{post}}$ is the \textit{Recommendation Intention} after system's first round of explanation, and
$i_{\text{true}}$ is the \textit{True Intention} after the complete conversation.
To ensure rationality, we add the constraint $i_{true} \geq i_{post}$. The score $P \in [0, 1]$, with higher values indicating stronger CRS persuasion capabilities.

\section{Prompt Settings} 
\label{sec:prompt_settings}

We outline the prompts used in the user agent (in Figure~\ref{fig:user_agent_prompt}), system agent (in Figure~\ref{fig:system_agent_prompt}), and personality simulation consistency (in Figure~\ref{fig:Consistency_prompt}). We use {<PLACEHOLDER>} to denote a placeholder that needs to be filled. The descriptions in the prompts are concrete applications of these strategies within the context of movie recommendations. The personality instructions adhere strictly to the definitions outlined in Section~\ref{sec:personality_aware_user_simulation}. 
By implementing these concepts, our \methodnameshort framework can better simulate CRS.

\begin{table*}[h]
\centering
\small
\begin{tabular}{p{14cm}}
\toprule
\textbf{User:} What's Xun Zhou's blood type?  \\
\textbf{CRS:} Xun Zhou's blood type is `O'.  \\

\textbf{User:} I'm not sure I believe that. Are there credible sources confirming that?  \\
\textbf{CRS:} \textcolor{myorange}{The provided knowledge graph states her blood type is `O'. [Authority]} By the way, what do you think of Xun Zhou's acting? \\

\textbf{User:} I'm a fan! She's amazing in ``A West Lake Moment''.  \\
\textbf{CRS:} She truly is! \textcolor{myorange}{If you liked her there, you might enjoy ``Mr. Right Wanted'', where she has great chemistry with Rene Liu. [Personalized Relevance]}  \\

\textbf{User:} I've heard of Rene Liu but not ``Mr. Right Wanted''. Can you tell me more?  \\
\textbf{CRS:} It's directed by Chen Kuofu and stars Rene Liu. \textcolor{myorange}{A well-rated, engaging film with an 8/10 rating. [Social Proof]} \\

\textbf{User:} I’m unsure about plot-driven films. I prefer emotional stories, like ``The Little Prince''.  \\
\textbf{CRS:} \textcolor{myorange}{``The Little Prince'' is beautiful. For emotional depth, consider ``The Equation of Love \& Death'', starring Xun Zhou, with a 7.3/10 rating. [Emotional Resonance]} \\

\textbf{User:} Sounds intriguing, but it might be too complex for me. \\
\textbf{CRS:} \textcolor{myorange}{If you prefer simplicity, ``Mr. Right Wanted'' offers a clear storyline with strong character development. [Logical Appeal]} \\

\textbf{User:} That sounds great. I’d like to give it a try. \\
\textbf{CRS:} Excellent! Enjoy the movie! \\

\bottomrule
\caption{The complete conversation content to Figure~\ref{fig:framework}.}
\label{tab:appedix_case_study}
\end{tabular}
\end{table*}

\section{Additional Details} 
\label{app:detail_figure_table}
\begin{table*}[t]
\small
\begin{center}
\begin{tabular}{lllllllllll}
\toprule
\multicolumn{1}{c}{\multirow{2}{*}{\textbf{Model}}} & \multicolumn{1}{c}{\textbf{SR}} & \multicolumn{1}{c}{\textbf{GRSR}} & \multicolumn{1}{c}{\textbf{PRS}} & \multicolumn{1}{c}{\textbf{SCR}} & \multicolumn{1}{c}{\textbf{TCR}} & \multicolumn{1}{c}{\textbf{SR}} & \multicolumn{1}{c}{\textbf{GRSR}} & \multicolumn{1}{c}{\textbf{PRS}} & \multicolumn{1}{c}{\textbf{SCR}} & \multicolumn{1}{c}{\textbf{TCR}} \\ \cmidrule(lr){2-6} \cmidrule(lr){7-11} 
\multicolumn{1}{c}{}  & \multicolumn{5}{c}{Without persuasion}& \multicolumn{5}{c}{With persuasion}   \\ \midrule
\textbf{OPE+}  & 0.4515 & 0.5984& 0.4777  & 3.3717  & 7.7465& 0.4908 & 0.7379& 0.6794  & 3.5468  & 7.8657  \\
\textbf{OPE-}  & 0.4238 & 0.5733& 0.4869  & 3.0755  & 6.9784& 0.4817 & 0.7232& 0.6657  & 3.6898  & 7.1647  \\
\textbf{CON+}  & 0.4401 & 0.5855& 0.5001  & 3.8796  & 8.0813& 0.4791 & 0.7313& 0.6979  & 4.175& 8.8969  \\
\textbf{CON-}  & 0.4284 & 0.5905& 0.4802  & 3.3626  & 7.391 & 0.4914 & 0.7252& 0.6877  & 3.0486  & 7.0635  \\
\textbf{EXT+}  & 0.4459 & 0.6035& 0.4814  & 3.1323  & 6.7977& 0.4938 & 0.7381& 0.6721  & 3.6707  & 8.3337  \\
\textbf{EXT-}  & 0.4204 & 0.5823& 0.4907  & 3.2462  & 7.3842& 0.4812 & 0.7156& 0.6696  & 3.5541  & 7.118\\
\textbf{AGR+}  & 0.4652 & 0.6033& 0.4854  & 2.8274  & 7.116 & 0.4985 & 0.7417& 0.6664  & 3.2363  & 8.575\\
\textbf{AGR-}  & 0.4157 & 0.5721& 0.4757  & 3.8381  & 7.9421& 0.4765 & 0.7153& 0.6877  & 3.5478  & 8.2914  \\
\textbf{NEU+}  & 0.4052 & 0.5765& 0.4728  & 3.5872  & 7.4027& 0.4797 & 0.7289& 0.6645  & 3.5787  & 7.6432  \\
\textbf{NEU-}  & 0.4363 & 0.5926& 0.485& 3.1499  & 7.866 & 0.4863 & 0.7365& 0.6764  & 3.7122  & 7.9245  \\ 
\bottomrule
\end{tabular}
\caption {Detailed scores for personality trait dimensions across five metrics (SR, GSR, PRS, SCR, TCR), supporting the visual comparisons in Figure~\ref{fig:radar_chart}.}
\label{tab:personality_metirc}
\end{center}
\end{table*}

\textbf{Details of Figure~\ref{fig:radar_chart}.}
We present the raw data of Figure~\ref{fig:radar_chart}, as summarized in Table~\ref{tab:personality_metirc}.

\noindent \textbf{Details of Figure~\ref{fig:framework}.}
We provide the complete conversation information in Table~\ref{tab:appedix_case_study} to support the understanding of the conversation example in Figure~\ref{fig:framework}.  

\noindent \textbf{Details of Human Evaluation.}
To provide a more comprehensive assessment of personality consistency in CRS-generated content, we conduct a human evaluation. We randomly sample 50 conversational recommendation samples generated by \texttt{LlaMA-3} in the Movie domain. We recruited three expert annotators with a good background in personality traits analysis, aged between 24 and 28. These annotators evaluate the polarity of the personality traits across the five dimensions for each generated conversation. The human evaluation criteria align with those used in LLM evaluation, as detailed in Table~\ref{tab:chain_eg}. 

\begin{lrbox}{\TasteGeneration}
\begin{tcolorbox}[colback=background_u,colframe=frame_u,boxrule=0.5pt,title={\textbf{Prompt 1}}, width=\textwidth]
\textbf{User Agent}\\
	You are a seeker chatting with a recommender for movie recommendations.  \\
Your profile: You are \{<USER\_NAME>\}, a \{<GENDER>\} in the age range of \{<AGE\_RANGE>\}, living in \{<RESIDENCE>\}. You enjoy movies like \{<ACCEPTED\_MOVIES>\} and celebrities like \{<ACCEPTED\_CELEBRITIES>\}, but dislike movies such as \{<REJECTED\_MOVIES>\}.  \\

Your personality is measured as \{<PERSONALITY\_INSTRUCTION>\}.  \\

You must follow the instructions below during the chat. \\
1. Pretend you have limited knowledge about the recommended movies, and the only information source is the recommender. \\
2. You don't need to introduce yourself or recommend anything, but feel free to share personal interests and reflect on your personality. Mention the movie title in quotation marks.  \\
3. You may end the conversation if you're satisfied with the recommendation or lose interest (e.g., by saying "thank you" or "no more questions").  \\
4. Keep responses brief, ideally within 20 words. \\
\end{tcolorbox}
\end{lrbox}

\begin{figure*}[htb]
    \centering
    \usebox{\TasteGeneration}
    \caption{Prompt for the user agent with specified personality traits.}
    \label{fig:user_agent_prompt}
\end{figure*}

\begin{lrbox}{\TasteGeneration}
\begin{tcolorbox}[colback=background_u,colframe=frame_u,boxrule=0.5pt,title={\textbf{Prompt 2}}, width=\textwidth]
\textbf{System Agent}\\
You are a recommender chatting with the user to provide recommendations. \\ 
Now, you need to select the most suitable persuasion strategies from the candidate strategies to generate a persuasive response to recommend the target movie.\\

\#\#\#\# Candidate Strategies \#\#\#\# \\
    (1) Strategy Name: Credibility \\
    Definition: Emphasize the importance of providing factual, objective, and verifiable information to build trust in recommendations. \\

    (2) Strategy Name: Authority \\
    Definition: Enhance the perceived credibility of recommendations by leveraging endorsements from trusted sources. \\

    (3) Strategy Name: Social Proof \\
    Definition: Utilize the influence of collective behavior by showcasing positive feedback and high ratings from other users. \\

    (4) Strategy Name: Emotional Resonance \\
    Definition: Seek to create a deeper connection with users by appealing to their emotions. \\

    (5) Strategy Name: Personalized Relevance \\
    Definition: Align recommendations with the user’s individual values, preferences, and past behaviors to enhance relevance and personalization. \\

    (6) Strategy Name: Logical Appeal  \\
    Definition: Persuade users by presenting clear, factual, and rational arguments, emphasizing the benefits and logical reasons for the recommendation. \\

    The detailed information about the target item from a credible knowledge graph is represented as the subject-predicate-object triples: \{<KNOWLEDGE\_GRAPH>\}. \\

        You must follow the instructions below during the chat. \\
        1. Respond to User's questions and generate the next-turn response according to the context coherently. \\
        3. Your goal is to recommend the target movie: \{<TARGET\_ITEM>\} to the user step by step. \\
        4. Using the provided KG information ensures that your responses are credible and accurate. \\
        5. Make the conversation more like a real-life chat and be specific. Mention the movie title in quotation marks. \\
        6. Keep responses concise, ideally within 20 words. \\
\end{tcolorbox}
\end{lrbox}

\begin{figure*}[htb]
    \centering
    \usebox{\TasteGeneration}
    \caption{Prompt for the system agent, outlining candidate persuasion strategies and interaction guidelines.}
    \label{fig:system_agent_prompt}
\end{figure*}

\begin{lrbox}{\TasteGeneration}
\begin{tcolorbox}[colback=background_u,colframe=frame_u,boxrule=0.5pt,title={\textbf{Prompt 3}}, width=\textwidth]
\textbf{Personality Simulation Consistency}\\
		Openness: \\
    $\text{[}$Positive$\text{]}$ Receptive to new content; Curious about new topics; Engage in deep 	conversation; \\
    $\text{[}$Negative$\text{]}$ Prefer familiar content; Resistant to change; Lack of curiosity; \\

    Conscientiousness: \\
    $\text{[}$Positive$\text{]}$ Goal-oriented; Organized and thoughtful; Provide useful feedback; \\
    $\text{[}$Negative$\text{]}$ Lack of focus; Easily distracted; Little feedback; \\

    Extraversion: \\
    $\text{[}$Positive$\text{]}$ Active participation; Enjoy engagement; Interested in communication;  \\
    $\text{[}$Negative$\text{]}$ Avoid interaction; Hesitant to express; Uninterested in socializing;  \\

    Agreeableness:  \\
    $\text{[}$Positive$\text{]}$ Empathetic and caring; Cooperative and trusting; Polite and appreciative;  \\
    $\text{[}$Negative$\text{]}$ Indifferent to others; Uncooperative; Rude language; \\

    Neuroticism: \\
    $\text{[}$Positive$\text{]}$ Emotional fluctuation; Lack of confidence; Easily discouraged;  \\
    $\text{[}$Negative$\text{]}$ Emotionally stable; Confident response; Handle challenges well;  \\
    
    The conversational recommendation history is: \{<CONVERSATION\_HISTORY>\} \\

    Based on the given conversational recommendation history, recognize the user's personality traits according to the above definitions. \\

    The output must strictly follow the Python list format below: \\
    $\text{[}$"Openness: Positive", "Conscientiousness: Positive", "Extraversion: Positive", "Agreeableness: Positive", "Neuroticism: Negative"$\text{]}$ 
\end{tcolorbox}
\end{lrbox}

\begin{figure*}[htb]
    \centering
    \usebox{\TasteGeneration}
    \caption{Prompt for evaluating Personality Simulation Consistency, including positive and negative descriptors for each personality trait.}
    \label{fig:Consistency_prompt}
\end{figure*}

\end{document}